\pdfoutput=1

\documentclass[11pt]{article}

 \usepackage[final]{acl}

\usepackage{times}
\usepackage{latexsym}

\usepackage[T1]{fontenc}

\usepackage[utf8]{inputenc}

\usepackage{microtype}

\usepackage{inconsolata}

\usepackage{graphicx}

\usepackage{enumitem}

\usepackage{printlen}

\usepackage{pgfplots}
\pgfplotsset{compat=1.17}

\usepackage{float}
\usepackage{hyperref}
\usepackage{url}
\usepackage{booktabs}
\usepackage{tabularray}

\usepackage[ruled,vlined]{algorithm2e}

\SetAlgorithmName{Algo}{}{}

\SetKwInput{KwIn}{Input}
\SetKwInput{KwOut}{Output}

\usepackage{longtable}

\usepackage{array}
\usepackage{pifont}
\usepackage{tabularx}
\usepackage{adjustbox}
\usepackage{multirow}
\usepackage{enumitem}
\usepackage{xspace}
\usepackage[breakable]{tcolorbox}
\usepackage{booktabs,amsfonts,dcolumn}
\usepackage{amsmath,amsthm,amsfonts,amssymb,bm,stmaryrd,bbm}
\usepackage{colortbl}
\usepackage{wrapfig}
\usepackage{placeins}
\usepackage{titlesec}
\usepackage{setspace}

\titlespacing{\paragraph}{
  0pt}{
  0.3\baselineskip}{
  0.5em}
\titlespacing*{\section}{0pt}{0.4\baselineskip}{0.4\baselineskip}
\titlespacing*{\subsection}{0pt}{0.3\baselineskip}{0.3\baselineskip}

\hypersetup{
    colorlinks,
    linkcolor={red!50!black},
    citecolor={blue!50!black},
    urlcolor={blue!80!black}
}
\usepackage{makecell}

\usepackage{cascadia-code}

\usepackage{courier}

\usepackage{cleveref}

\usepackage{helvet}

\usepackage{subcaption}
\usepackage{tikz}
\usetikzlibrary{calc,quotes}
\usetikzlibrary{positioning,shapes.misc}

\pgfplotsset{colormap/jet}
\usepackage{pifont}

\usepackage{tcolorbox}
\tcbuselibrary{skins}

\newcommand{\green}[1]{\textcolor[RGB]{74,103,65}{#1}}

\newcommand{\ourdataset}{\textsc{PairJudge-432K}\xspace}
\newcommand{\ourrms}{{PairJudge RMs}\xspace}
\newcommand{\ourrm}{{PairJudge RM}\xspace}

\newcommand{\criticmodel}{Critic Model\xspace}
\newcommand{\git}{\raisebox{-1.5pt}{\includegraphics[height=1.05em]{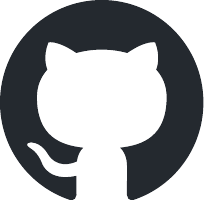}}\xspace}
\newcommand{\hf}{\raisebox{-1.5pt}{\includegraphics[height=1.05em]{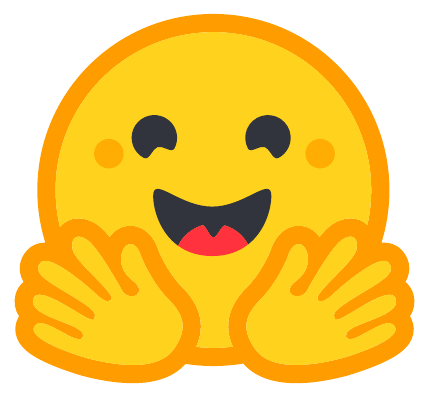}}\xspace}

\title{PairJudge RM: Perform Best-of-N Sampling with Knockout Tournament}

\author{
  Yantao Liu$^{1}$,~~Zijun Yao$^{2}$,~~Rui Min$^{3}$,~~Yixin Cao$^1$,~~Lei Hou$^2$,~~Juanzi Li$^2$\\
  $^1$Fudan University, $^2$Tsinghua University, $^3$Hong Kong University of Science and Technology\\
  \texttt{ricardoliu@outlook.com, yaozj20@mails.tsinghua.edu.cn} \\
  \vspace{0.1cm}
  \normalsize
  \git Code: \url{https://github.com/THU-KEG/PairJudgeRM/} \\
  \normalsize
  \hf Model: \url{https://huggingface.co/THU-KEG/PairJudgeRM} \\
  \normalsize
  \hf Dataset: \url{https://huggingface.co/datasets/THU-KEG/PairJudge-432K} \\
}

\begin{document}
\maketitle

\NewTotalTColorBox[auto counter]{\pairwise}{ +m +m }{ 
    notitle,
    colback=blue!5!white,
    colbacklower=white,
    frame hidden,
    boxrule=0pt,
    width=11cm,
    bicolor,
    sharp corners,
    borderline west={4pt}{0pt}{blue!50!black},
    fontupper=\sffamily
}{
    \textcolor{blue!50!black}{
        \sffamily
        \textbf{}%
    }%
    #1
    \tcblower%
    \textcolor{blue!50!black}{
        \sffamily
        \textbf{}%
    }%
    #2
}

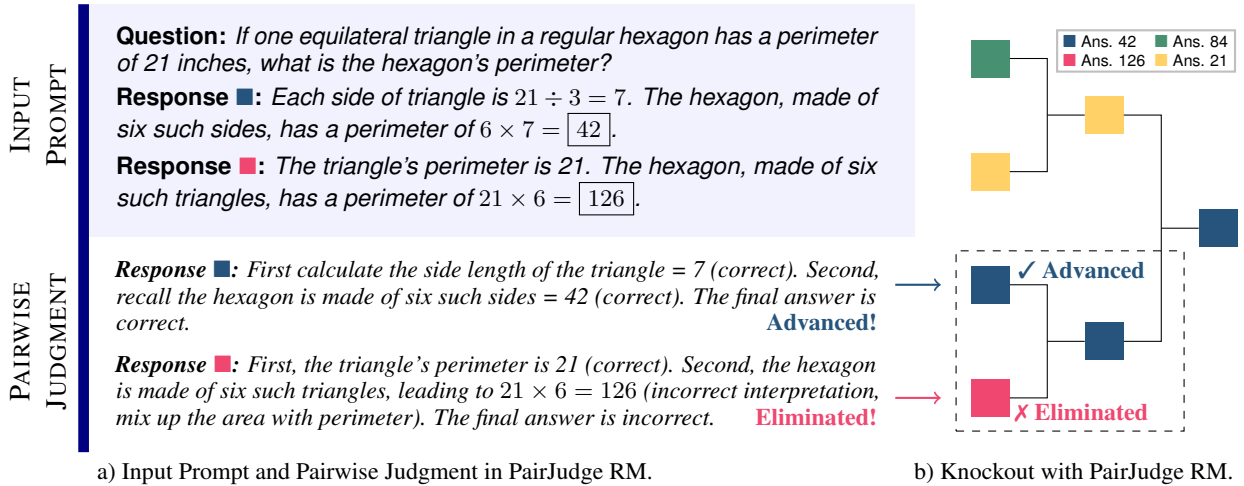
\begin{figure*}[h]
  \definecolor{C1}{HTML}{26547C}
  \definecolor{C3}{HTML}{ef476f}
  \definecolor{C2}{HTML}{ffd166}
  % forest green
  \definecolor{C4}{HTML}{479071}
  
  \hskip-0.3cm
  \begin{tikzpicture}
    \begin{scope}[xshift=6.5cm,yshift=+2.25cm,every node/.style={rectangle,minimum size=0.5cm}]

      \node[fill=C4] (x1) at (0, 0) {};
      \node[fill=C2] (x2) at ($(x1)+(0, -1.5)$) {};
      \node[fill=C1] (x3) at ($(x2)+(0, -1.5)$) {};
      \node[fill=C3] (x4) at ($(x3)+(0, -1.5)$) {};
      
      \node[] at ($(x3)+(1.2,0.2)$) {\color{C1}\small \ding{51} \bf Advanced};

      \node[] at ($(x4)+(1.2,-0.2)$) {\color{C3}\small \ding{55} \bf Eliminated};
      \node[fill=C2] (y1) at ($(x1)!0.5!(x2)+(1.5, 0)$) {};
      \node[fill=C1] (y2) at ($(x3)!0.5!(x4)+(1.5, 0)$) {};

      \node[fill=C1] (z1) at ($(y1)!0.5!(y2)+(1.5, 0)$) {};

      \node[rectangle,anchor=north west,fill=none,draw,dashed,minimum width=3cm,minimum height=68] at ($(x3.north west)+(-0.2,0.2)$) (card) {};

      \foreach \x/\y/\z in {x1/x2/y1, x3/x4/y2, y1/y2/z1}{
          \draw (\x) -- ($(\x)+(0.75, 0)$) -- ($(\y)+(0.75, 0)$) -- (\y);
          \draw ($(\x)!0.5!(\y)+(0.75, 0)$) -- (\z);
        }
    \end{scope}

    \node (x1) at (0,0) {\begin{tcolorbox}[
      notitle,
      colback=blue!5!white,
      colbacklower=white,
      frame hidden,
      boxrule=0pt,
      width=11cm,
      bicolor,
      sharp corners,
      borderline west={4pt}{0pt}{blue!50!black},
      fontupper=\sffamily
      ]
      \small
      \textbf{Question:} \textit{If one equilateral triangle in a regular hexagon has a perimeter of 21 inches, what is the hexagon’s perimeter?}
      
      \vspace{0.1cm}
      \textbf{Response \textcolor{C1}{$\blacksquare$}:} \textit{Each side of triangle is $21\div3=7$. The hexagon, made of six such sides, has a perimeter of $6\times7=\boxed{42}$.}
      
      \vspace{0.1cm}
      \textbf{Response \textcolor{C3}{$\blacksquare$}:} \textit{The triangle’s perimeter is 21. The hexagon, made of six such triangles, has a perimeter of $21\times6=\boxed{126}$.}
      
      \tcblower
      
      \small
      \textit{\textbf{Response \textcolor{C1}{$\blacksquare$}:}} \textit{First calculate the side length of the triangle = 7 (correct). Second, recall the hexagon is made of six such sides = 42 (correct). The final answer is correct.}\hfill\textcolor{C1}{\textbf{Advanced!}}
      
      \vspace{0.2cm}
      \textit{\textbf{Response \textcolor{C3}{$\blacksquare$}:}} \textit{First, the triangle's perimeter is 21 (correct). Second, the hexagon is made of six such triangles, leading to $21\times6=126$ (incorrect interpretation, mix up the area with perimeter). The final answer is incorrect.}\hfill\textcolor{C3}{\textbf{Eliminated!}}
      
  \end{tcolorbox}};

  \node[rotate=90] at (-6,1.5) {\makecell[c]{\scshape Input \\ \scshape Prompt}};

  \node[rotate=90] at (-6,-1.5) {\makecell[c]{\scshape Pairwise \\ \scshape Judgment}};

  % add a arrow
  \draw[->,thick,C1] (5.25,-0.75) -- (5.85,-0.75);
  \draw[->,thick,C3] (5.25,-2.25) -- (5.85,-2.25);

  \node[rectangle, thick, lightgray, draw, inner sep=2pt] at (8.55,2.35) {
  \tiny
  \begin{minipage}[c]{2.2cm} % Adjusted width for the 2x2 grid
    \sffamily
    \textcolor{C1}{$\blacksquare$} \textcolor{black}{Ans. 42\phantom{0}} \hspace{0.00cm} 
    \textcolor{C4}{$\blacksquare$} \textcolor{black}{Ans. 84}\\
    \textcolor{C3}{$\blacksquare$} \textcolor{black}{Ans. 126} \hspace{0.00cm} 
    \textcolor{C2}{$\blacksquare$} \textcolor{black}{Ans. 21}
  \end{minipage}
};

  \node[rectangle,inner sep=3pt] at (-0.25,-3.25) {
  \small
    \begin{minipage}[c]{10cm}
    % use as subcaption
    a) Input Prompt and Pairwise Judgment in \ourrm.
    \end{minipage}
  };

  \node[rectangle,inner sep=3pt] at (8.00,-3.25) {
  \small
    \begin{minipage}[c]{5cm}
    % use as subcaption
    b) Knockout with \ourrm.
    \end{minipage}
  };

  \end{tikzpicture}
  \caption{
  An example of the knockout tournament using the \ourrm. Panel (a) shows a pairwise judgment in the knockout tournament (Panel b) between response \textcolor{C1}{$\blacksquare$} and \textcolor{C3}{$\blacksquare$}.
  The \ourrm takes a question and responses \textcolor{C1}{$\blacksquare$}, \textcolor{C3}{$\blacksquare$} as input, then judges their correctness pairwise.
  In the judgment, the \ourrm correctly identifies the response \textcolor{C1}{$\blacksquare$} as correct and response \textcolor{C3}{$\blacksquare$}  as incorrect, leading to the elimination of the \textcolor{C3}{$\blacksquare$}.
  The pairwise judgment process continues iteratively in the knockout tournament until only one response remains.
  For this question, the response \textcolor{C1}{$\blacksquare$} is selected as the best candidate solution and Answer $42$ is the final answer through the knockout.
  }

  \label{fig:pairwise-knockout}
\end{figure*}

\begin{abstract}
    Best-of-N (BoN) sampling, a common strategy for test-time scaling of Large Language Models (LLMs), relies on reward models to select the best candidate solution from multiple generations. 
    However, traditional reward models often assign arbitrary and inconsistent scores, limiting their effectiveness. 
    To address this, we propose a Pairwise Judge Reward Model (\ourrm) combined with a knockout tournament for BoN sampling. 
    Instead of assigning absolute scores, given one math problem, \ourrm judges two candidate solutions' correctness with chain-of-thought reasoning simultaneously.
    This approach eliminates the need for scoring and enables cross-validation of solutions through parallel judgment. 
    In the knockout tournament, \ourrm conducts pairwise Judgment between candidate solutions and eliminates the incorrect ones iteratively.
    We construct \ourdataset, a large-scale dataset of 432K pairwise judgments derived from NumiaMath and annotated using \texttt{gemini-1.5-flash}, and train the \ourrm via supervised fine-tuning. 
    Experiments on MATH-500 and the Olympiad Bench demonstrate significant improvements over baseline reward models.
    And a 40\% to 60\% relative improvement is achieved on the top 50\% challenging problems.

\end{abstract}

\section{Introduction}

Recently, test-time scaling has garnered significant attention from the research community, as it draw a blueprint for the next stage of scaling of Large Language Models (LLMs)~\citep{snell2024scaling, wu2024empirical, openai_o1_preview}.
One of the most common practice to achieve test-time scaling is to use reward models (RMs) to perform the Best-of-N (BoN) Sampling at test time~\citep{wang2023math, lightman2023let, wang2024openr, zhang2024generative, yuan2024free}: the LLM generates $N$ candidate solutions for a given problem, and a learned reward model, scoring each candidate solution, selects the best one as the final output.
The effectiveness of this strategy hinges on how accurate the score assigned by the reward model is to the candidate solutions.

However, assigning accurate and consistent scores is inherently challenging, even for human experts~\citep{jonsson2007use,abdul2014study}. 
An experiment conducted in NeurIPS 2021 shows that for different human experts guided by the same rubric, the scores assigned to the same candidate paper can vary significantly~\citep{neurips2021}.
This limitation is particularly pronounced in reward models, which are typically trained to assign relative scores rather than absolute, meaningful scores~\citep{lambert2024rewardbench,liu2024rm}. 
As a result, the scores assigned by reward models are often arbitrary and inconsistent, hindering the performance of BoN sampling~\citep{liu2024rm,kim2024evaluating}.

To address this limitation, we propose a Pairwise Judge Reward Model (\ourrm) combined with a knockout tournament for BoN sampling. 
Instead of assigning absolute scores, \ourrm judges two candidate solutions' correctness simultaneously for a given reasoning problem.

In this setting, our approach eliminates the need for arbitrary scoring and enables cross-validation of solutions through pairwise judgment. 
To perform BoN sampling, we organize candidate solutions into a knockout tournament, where each pairwise judgment acts as a match. 
Rounds of matches are played until only one candidate remains, which is selected as the final output.
Figure~\ref{fig:pairwise-knockout} presents the knockout tournament process with our \ourrm.
Specifically, inlining with existing work~\citep{lightman2023let,wang2023math,snell2024scaling,wu2024empirical}, math reasoning tasks are used as the testbed to evaluate the BoN sampling performance.

We construct \ourdataset, a large-scale dataset of 432K pairwise judgments derived from NumiaMath~\citep{numina_math_datasets} and annotated using \texttt{gemini-1.5-flash}. 
Using this dataset, we train the \ourrm via supervised fine-tuning. 
Experiments on MATH-500 and the Olympiad Bench demonstrate that \ourrm significantly outperforms traditional discriminative reward models. 
On the top 50\% challenging problems in MATH-500, \ourrm achieves a 40\% to 60\% relative improvement over the baseline. 
Furthermore, our method outperforms the recently proposed \criticmodel~\citep{gao2024critics,mcaleese2024critics,zhang2024generative} under the same computational budget.

In summary, our contributions are as follows:
\begin{itemize}[noitemsep,topsep=0pt,parsep=0pt,partopsep=0pt]
    \item We propose a Pairwise Judge Reward Model (\ourrm) combined with a knockout tournament for BoN sampling. 
    This approach avoids the limitations of arbitrary scoring in traditional reward models and enables cross-validation of candidate solutions.
    \item We release \ourdataset, a large-scale dataset for training \ourrms containing 432K annotated pairwise judgments, along with its construction pipeline. 
    \item Experiments on MATH-500 and the Olympiad Bench demonstrate significant improvements compared to baselines in BoN sampling.
    Specifically, on the top 50\% difficult problems, \ourrm achieves a 40\% to 60\% relative improvement over baseline models.
\end{itemize}

\section{Preliminaries}

\paragraph{Best-of-N Sampling in Math Reasoning}
Given a math problem $x \in \mathcal{X}$ and the $N$ candidate solutions $\{y_1, y_2, \ldots, y_N\}$ sampled from a Large Language Model (LLM), the BoN Sampling aims to select the best candidate solution $y^*$ from the $N$ candidate solutions based on an external selection mechanism.
Typically, there are two types of reward models (RMs) serving as the external selection mechanism: the Outcome Reward Model and the Process Reward Model.

\paragraph{Outcome Reward Model}
Given a math problem $x$ and a candidate solution $y$, the Outcome Reward Model assigns a numerical score $s(y)$ to the candidate solution $y$.
The Outcome Reward Model selects the candidate solution with the highest score as the final output:
\begin{equation}
    y^* = \arg\max_{y \in \{y_1, y_2, \ldots, y_N\}} s(y).
    \label{eq:best-of-n}
\end{equation}

The Outcome Reward Model is typically trained on a preference dataset $\mathcal{D}$, consisting of pairs $(x, y_c, y_r)$, where $y_c$ is the chosen response and $y_r$ is the rejected response.
The model is trained to assign a higher reward to $y_c$ than to $y_r$, minimizing the following objective:
\begin{equation}
    \mathcal{L} = -\mathbb{E}\left[\log\sigma(R_{\psi}(x, y_c) - R_{\psi}(x, y_r))\right],
\end{equation}
where $\sigma(\cdot)$ is the sigmoid function and $\psi$ is the parameters of the  Reward Model.
$\mathcal{L}$ is the loss function for preference learning indicating the probability of the chosen response $y_c$ being preferred over the rejected response $y_r$.
This objective ensures that the reward model learns to identify responses that align better with human preferences.

\paragraph{Process Reward Model}
Given a math problem $x$ and a corresponding candidate solution $y$, the Process Reward Model first requires to split the candidate solution $y$ into a sequence of reasoning steps $\{a_1, a_2, \ldots, a_M\}$.
The Process Reward Model assigns a numerical score $s(a_i)$ to each reasoning step $a_i$.
The score of the entire candidate solution $y$ is the mean of the scores of all reasoning steps:
\begin{equation}
    s(y) = \frac{1}{M}\sum_{i=1}^{M}s(a_i).
\end{equation}
The Process Reward Model selects the candidate solution with the highest score as the final output with the same mechanism as the Outcome Reward Model in Equation~\ref{eq:best-of-n}.

The Process Reward Model is typically trained on a dataset with process labels $\mathcal{D}_{\text{proc}}$, where each solution $y$ to a problem $x$, the dataset contains a series of process labels $\{l_1, l_2, \ldots, l_M\}$, where $l_i \in \{0, 1\}$ indicates whether the reasoning step $a_i$ is correct or incorrect. Then the Process Reward Model is trained to predict the correctness of each reasoning step $a_i$.

\section{\ourrm and Knockout}

In this section, we introduce the \ourrm and the knockout tournament, which are the core components of our proposed method for performing BoN Sampling at test time.

\subsection{Pairwise Judge Reward Model}
\label{sec:pairwise-reward-model}

\paragraph{Definition}
Given a math problem $x$ and two candidate solutions $y_1$ and $y_2$, the \ourrm is designed to simultaneously check the correctness of the two candidate solutions. 
Specifically, the \ourrm is trained to judge the correctness of the two candidate solutions, denoted as $c_1$ and $c_2$, respectively.
\begin{equation}
    c_1, c_2 = \text{PairJudge}(x, y_1, y_2),
\end{equation}
where $c_1, c_2 \in \{0, 1\}$ indicate whether the candidate solutions $y_1$ and $y_2$ are correct or incorrect.

\paragraph{Implementation}
Inspired by the Generative Reward Model (GenRM)~\citep{zhang2024generative} and LLM-as-a-Judge~\citep{zheng2023judging}, we implement the \ourrm as a generative model. 
Specifically, given a math problem $x$ and two candidate solutions $y_1$ and $y_2$, the \ourrm first generates a reasoning text using chain-of-thought~\citep{wei2022chain} to verify the correctness of the two candidate solutions. 
Based on the reasoning text, the \ourrm then predicts the correctness of the two candidate solutions by directly generating the correctness labels $c_1$ and $c_2$. 
The detailed prompt for performing pairwise verification with chain-of-thought is provided in Table~\ref{tab:pairwise-prompt} in the Appendix.

\subsection{Knockout Tournament}

To perform BoN Sampling with the \ourrm, we introduce a knockout tournament to select the best candidate solution.

\begin{algorithm}
    \caption{Knockout for Best-of-N Sampling}
    \label{alg:knockout-tournament}
    \KwIn{
        Math problem $x$, \\
        $N$ candidate solutions $\mathcal{Y} = \{y_1, y_2, \dots, y_N\}$, \\
        Pairwise Judge Reward Model: $\text{PairJudge}$
    }
    \KwOut{Best candidate solution $y_{\text{best}}$}
    
    \BlankLine
    \textbf{Step 1: Group candidates into teams} \\
    Partition $\mathcal{Y}$ into $k$ teams, where members of a team share the same final answer.

    \BlankLine
    \textbf{Step 2: Initialize the knockout pool} \\
    Add all $N$ candidates to the initial pool $\mathcal{P}$.
    
    \BlankLine
    \textbf{Step 3: Perform the knockout rounds} \\
    \While{$|\mathcal{P}| > 1$ }{
        Pair each candidate $y_i$ with an unpaired $y_j$ from a different team. \\
        Remove $y_i$ and $y_j$ from $\mathcal{P}$. \\
        \ForEach{pair $(y_i, y_j)$}{
            Compute correctness scores $c_i, c_j$ using $\text{PairJudge}(x, y_i, y_j)$. \\
            \uIf{$c_i > c_j$}{$y_i$ advances.}
            \uElseIf{$c_j > c_i$}{$y_j$ advances.}
            \uElseIf{$c_i, c_j$ both correct}{
                Randomly select one to advance.
            }
            \Else{
                Both incorrect and eliminated.
            }
        }
        Add advancing candidates back to $\mathcal{P}$.
    }
    
    \BlankLine
    \textbf{Step 4: Return the best solution} \\
    Output the last remaining $y$ in $\mathcal{P}$ as $y_{\text{best}}$.
\end{algorithm}

Specifically, we first group the $N$ candidate solutions into $k$ teams, where candidates that share the same answer are placed in the same team. 
Then, we pair up the candidate solutions from each team to compete with candidate solutions from other teams. 
In each match, only the candidate solution that receives the correct label from the \ourrm advances to the next round. 
If both candidate solutions receive the correct label, one is randomly selected to advance. 
This process continues until only one candidate solution remains or early termination occurs when all candidate solutions are from the same team.

The detailed procedure of the knockout tournament is shown in Algorithm~\ref{alg:knockout-tournament}.

\section{\ourdataset dataset collection}
\label{sec:dataset}
To train the \ourrm, we collect a large-scale dataset named \ourdataset, 
which contains 432K annotated pairwise judgments derived from NumiaMath~\citep{numina_math_datasets} with \texttt{gemini-1.5-flash}.
In the following, we describe the detailed procedure of collecting the \ourdataset dataset.

\subsection{Dataset Format}
Since the \ourrm is designed as a generative model to judge the correctness of candidate solutions, the training dataset has the same format as the one for Supervised Fine-Tuning, consisting of prompt-completion pairs.
Specifically, each prompt is constructed by filling the template shown in Table~\ref{tab:pairwise-prompt} with a math problem $x$ and two candidate solutions $y_1$ and $y_2$.
The completion is a chain-of-thought reasoning text that judges the correctness of the two solutions and provides the correctness labels $c_1$ and $c_2$.

\subsection{Math Problem Collection}
We first collect math problems from the NumiaMath dataset~\citep{numina_math_datasets}, which contains 860K problems ranging from high school math exercises and international mathematics olympiad competition problems.
Because these data are primarily collected from online exam paper PDFs and mathematics discussion forums, we remove low-quality problems with messy formatting, OCR errors, or missing answers.
We also remove multiple-choice (MCQ) and True/False questions to avoid random guessing in candidate solutions.
Following community conventions, we remove proof problems as well, due to the difficulty of verifying candidate solutions.
The detailed filtering criteria are listed in Table~\ref{tab:math-problem-filter} of the Appendix.

\begin{table}[t]
    \centering
    \small
    \begin{tabular}{lrr}
        \toprule
        \textbf{Dataset} & \textbf{Original Count} & \textbf{Filtered Count} \\
        \midrule
        AMC/AIME         & 4,070                   & 289                      \\
        AoPS Forum       & 30,192                  & 9,017                    \\
        Chinese K-12     & 276,554                 & 63,779                   \\
        GSM8K            & 7,342                   & 6,539                    \\
        Math             & 7,477                   & 5,988                    \\
        Olympiads        & 150,563                 & 52,766                   \\
        ORCA Math        & 153,314                 & 149,550                  \\
        Synthetic AMC    & 62,108                  & 94                       \\
        Synthetic Math   & 167,874                 & 136,921                  \\
        \midrule
        \textbf{Total}   & \textbf{859,494}        & \textbf{425,943}         \\
        \bottomrule
    \end{tabular}
    \caption{Statistics of the datasets before and after filtering. 
    AMC-related datasets shrink significantly because most AMC problems are multiple-choice.}
    \label{tab:dataset-statistics}
    \vspace*{-0.1cm}
\end{table}

\subsection{Candidate Solution Generation}
For each math problem $x$, we generate $k=24$ candidate solutions $\{y_1, y_2, \ldots, y_k\}$ using {Llama-3.1-8B-instruct}~\citep{meta2025llama31}.
We employ the same four-shot in-context examples for all problems as the prompt.
The candidate solutions are decoded with a temperature of $1.0$ and a Top-P value of $0.5$ to balance diversity and quality.

\subsection{Pairwise Judgment Annotation}
We use \texttt{gemini-1.5-flash} to annotate the \ourrm training data on the NumiaMath dataset.
To align the generated training data distribution with the solution-judgment distribution in the knockout tournament, we conduct a knockout tournament for each math problem $x$ and its candidate solutions $\{y_1,y_2, \ldots, y_k\}$ to select the best solution $y_{\text{best}}$.
During the knockout tournament, we record all pairwise judgments among candidate solutions and retain only those records that correctly judge solution correctness for the \ourrm.
Specifically, due to cost considerations, we only run the knockout tournament for questions whose candidate solutions are not all correct or all incorrect.
As a result, we conducted 343K tournaments and recorded 2.2M comparisons.
Among these, 1.3M correctly evaluated both candidate solutions and were used as raw training data for the \ourrm.
Finally, we filtered out samples where the response did not strictly follow the instructions in Table~\ref{tab:pairwise-prompt}, ending up with 432K training samples.

\section{Experiments}
\begin{table*}[t]
    \centering
    \label{tab:refined-exp}
    \resizebox{\textwidth}{!}{
    \begin{tabular}{@{}l|l|ccccccccc|c@{}}
    \toprule
    \multirow{2}{*}{\bf Type} & \multirow{2}{*}{\bf Reward Model} & \multicolumn{3}{c|}{\bf Llama-3.1-8B-Inst} & \multicolumn{3}{c|}{\bf Qwen-2.5-7B-Inst} & \multicolumn{3}{c|}{\bf Llama-3.1-70B-Inst} & \multirow{2}{*}{\bf Avg.} \\ \cmidrule(lr){3-11}
    & & @16 & @32 & \multicolumn{1}{c|}{@64} & @16 & @32 & \multicolumn{1}{c|}{@64} & @16 & @32 & @64 & \\ \midrule
    \multicolumn{12}{c}{\it MATH-500} \\ \midrule
    \multirow{4}{*}{\bf ORM}
    & ArmoRM-Llama3-8B & 51.6 & 49.2 & \multicolumn{1}{c|}{49.8} & \underline{77.6} & 77.4 & \multicolumn{1}{c|}{76.4} & 64.8 & 64.8 & 65.8 & 64.2 \\ 
    & SkyworkRM-Llama3.1-8B & 51.4 & 51.0 & \multicolumn{1}{c|}{51.0} & \underline{77.6} & 76.4 & \multicolumn{1}{c|}{\underline{78.0}} & 66.4 & 66.6 & 67.4 & 65.1 \\
    & EurusRM-7B & 55.2 & 53.4 & \multicolumn{1}{c|}{53.4} & 76.6 & 77.0 & \multicolumn{1}{c|}{77.4} & 68.0 & 66.6 & 67.6 & 66.1 \\
    & Pair-Preference-Llama3-8B & 48.0 & 47.6 & \multicolumn{1}{c|}{49.0} & 76.0 & 77.4 & \multicolumn{1}{c|}{75.6} & 64.0 & 63.4 & 60.2 & 62.4 \\\midrule
    \multirow{4}{*}{\bf PRM}
    
    & Math-Shepherd-7B & 49.5 & 50.1 & \multicolumn{1}{c|}{49.2} & 74.7 & 75.3 & \multicolumn{1}{c|}{75.9} & 63.5 & 62.8 & 63.6 & 62.7 \\
    & RLHFlow-8B-Mistral-Data & 51.0 & 51.0 & \multicolumn{1}{c|}{50.2} & 75.4 & 76.2 & \multicolumn{1}{c|}{76.6} & 64.0 & 63.0 & 64.8 & 63.6 \\
    & RLHFlow-8B-DS-Data & 55.2 & 57.0 & \multicolumn{1}{c|}{56.2} & 75.8 & 76.0 & \multicolumn{1}{c|}{76.2} & 66.2 & 66.4 & 65.4 & 66.0 \\
    
    & RLHFlow-8B-LLaMA-Data & 55.5 & 56.8 & \multicolumn{1}{c|}{56.0} & 76.0 & 76.3 & \multicolumn{1}{c|}{76.5} & 66.7 & 67.0 & 66.0 & 66.3 \\\midrule
    & Majority Voting & \underline{57.0} & \underline{58.8} & \multicolumn{1}{c|}{\underline{58.8}} & {77.4} & \underline{77.6} & \multicolumn{1}{c|}{\underline{78.0}} & \underline{70.2} & \underline{72.8} & \underline{73.6} & \underline{69.4} \\\midrule
    & \ourrm \& Knockout & \textbf{61.0} & \textbf{64.6} & \multicolumn{1}{c|}{\textbf{65.6}} & \textbf{80.2} & \textbf{79.8} & \multicolumn{1}{c|}{\textbf{80.4}} & \textbf{72.2} & \textbf{75.6} & \textbf{77.4} & \textbf{73.0} \\\midrule
    \multicolumn{12}{c}{\it Olympiad Bench} \\ \midrule
    \multirow{4}{*}{\bf ORM}
    & ArmoRM-Llama3-8B & 16.1 & 15.9 & \multicolumn{1}{c|}{16.7} & 39.3 & 40.1 & \multicolumn{1}{c|}{40.4} & 29.2 & 29.8 & 30.1 & 28.7 \\ 
    & SkyworkRM-Llama3.1-8B & 19.9 & 20.0 & \multicolumn{1}{c|}{18.7} & 39.9 & {40.0} & \multicolumn{1}{c|}{\underline{41.0}} & 29.8 & 30.4 & 29.8 & 29.4 \\
    & EurusRM-7B & \underline{20.4} & 19.6 & \multicolumn{1}{c|}{20.1} & 37.9 & 39.4 & \multicolumn{1}{c|}{39.1} & 30.1 & 30.7 & 32.4 & 30.0 \\
    & Pair-Preference-Llama3-8B & 17.7 & 19.1 & \multicolumn{1}{c|}{17.2} & 39.4 & 38.9 & \multicolumn{1}{c|}{38.1} & 26.5 & 27.4 & 25.5 & 27.8 \\\midrule
    \multirow{4}{*}{\bf PRM}
    & Math-Shepherd-7B & 15.2 & 13.7 & \multicolumn{1}{c|}{13.1} & 34.8 & 34.5 & \multicolumn{1}{c|}{35.1} & 25.3 & 26.0 & 24.1 & 24.6 \\
    & RLHFlow-8B-Mistral-Data & 16.4 & 14.5 & \multicolumn{1}{c|}{14.5} & 36.1 & 35.9 & \multicolumn{1}{c|}{36.3} & 26.7 & 27.1 & 25.2 & 25.9 \\
    & RLHFlow-8B-DS-Data & 18.5 & 19.6 & \multicolumn{1}{c|}{19.3} & 35.4 & 34.8 & \multicolumn{1}{c|}{34.2} & 28.9 & 29.5 & 30.1 & 27.8 \\
    
    & RLHFlow-8B-LLaMA-Data & 18.7 & 20.0 & \multicolumn{1}{c|}{19.7} & 35.8 & 35.2 & \multicolumn{1}{c|}{34.7} & 29.1 & 29.4 & 30.3 & 28.1 \\\midrule
    & Majority Voting & {20.3} & \underline{22.4} & \multicolumn{1}{c|}{\underline{23.3}} & \underline{40.0} & \bf{40.7} & \multicolumn{1}{c|}{{39.9}} & \underline{35.6} & \underline{35.9} & \underline{36.7} & \underline{32.8} \\\midrule
    & \ourrm \& Knockout & \textbf{22.7} & \textbf{24.9} & \multicolumn{1}{c|}{\textbf{25.5}} & \textbf{41.9} & \underline{40.2} & \multicolumn{1}{c|}{\textbf{41.2}} & \textbf{33.9} & \textbf{36.7} & \textbf{37.8} & \textbf{33.9} \\\bottomrule
    \end{tabular}
    }
    \caption{Different reward models' best-of-N sampling performance on MATH-500 and Olympiad Bench with three different LLMs: Llama-3.1-8B-Inst, Qwen-2.5-7B-Inst, and Llama-3.1-70B-Inst. The results are reported in terms of accuracy.
    The pass@1 accuracy of these three LLMs are {42.0, 73.6, and 59.2} on MATH-500, and {12.3, 35.7, and 25.9} on Olympiad Bench, respectively.
    @16, @32, and @64 denote the accuracy with Best-of-16, Best-of-32, and Best-of-64 sampling, respectively.
    The best results are in bold, and the second-best results are underlined.}
    \label{tab:main-exp}
\end{table*}
In this section, we demonstrate the effectiveness of the \ourrm and the knockout tournament in performing BoN Sampling at test time.
We first introduce the experimental setup, including the dataset, evaluation metrics, and baselines.
Then, we present the experimental results and analysis.

\begin{figure*}[t]
    \centering
    \includegraphics[width=\textwidth]{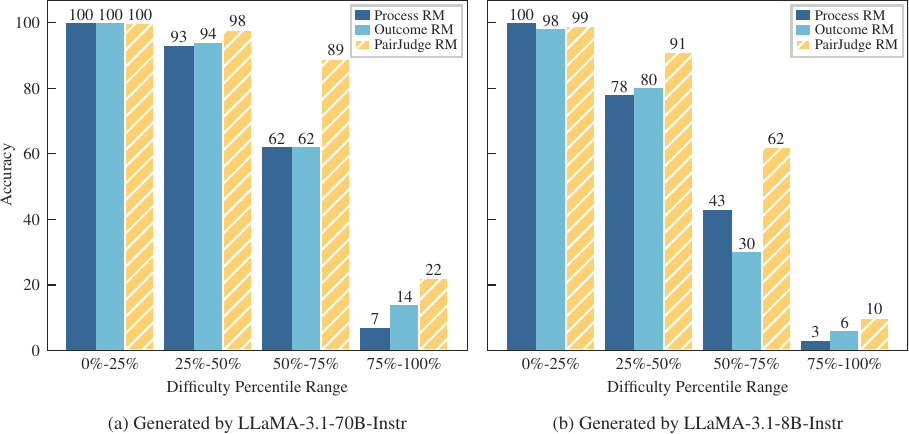}
    \caption{Comparison of Process RM, Outcome RM, and \ourrm across difficulty percentiles in MATH-500. Candidate solutions are generated by (a) LLaMA-3.1-70B-Instr and (b) LLaMA-3.1-8B-Instr. Process RM and Outcome RM refer to EurusRM-7B and RLHFlow-8B-DS-Data, respectively. As shown, \ourrm consistently outperforms both, except on the easiest problems. Notably, for the hardest 50\% problems, \ourrm achieves a 40\%–60\% relative improvement.}
    \label{fig:difficulty-exp}
\end{figure*}

\subsection{Experimental Setup}
\label{sec:exp-setup}
\paragraph{Dataset}
We evaluate BoN Sampling on MATH-500~\cite{hendrycksmath2021} and Olympiad Bench~\cite{he-etal-2024-olympiadbench} to coverage from the high-school-level math problems to the olympiad-level math problems.
To study the generalizability of our \ourrm, we test it with three LLMs that have different capabilities and come from different model families: {Llama-3.1-8B-Instruct}~\citep{llama3modelcard}, {Llama-3.1-70B-Instruct}~\citep{llama3modelcard}, and {Qwen2.5-7B-Instruct}~\citep{yang2024qwen2}.

\paragraph{Training Details}
We use {Qwen2.5-7B-Instruct} as the base model and perform supervised fine-tuning on our \ourdataset dataset to obtain the \ourrm.
We set the learning rate to $1\times 10^{-5}$ with the Adam optimizer and a batch size of 128.
The model is trained for $8$ epochs.

\paragraph{Baselines}
We compare our \ourrm with both outcome and process reward model, which is trained to assign a score to each candidate solution and then select the candidate solution with the highest score as the final output.
For the Outcome Reward Model, we use EurusRM-7B~\cite{yuan2024advancing}, SkyworkRM-Llama3.1-8B~\cite{skyworkreward2024}, and ArmoRM-Llama3-8B~\cite{gao2024interpretable} as representatives of state-of-the-art outcome reward models.
In Outcome Reward Model, we also include a Pair-Preference-Llama3-8B as the representative of the pairwise preference model~\cite{dong2024rlhf,zhao2023slic,ye2024online,llm-blender-2023,lee-etal-2024-aligning}.
These models also take the candidate solution pairs as input, but they are trained as same as the outcome reward model to assign scores to the candidate solutions, instead of judging the correctness of the candidate solutions via chain-of-thought reasoning.
For the Process Reward Model, we leverage three off-the-shelf open-source models: Math-Shepherd~\citep{wang2023math}, RLHFlow-8B-Mistral-Data, and RLHFlow-8B-Deepseek-Data~\citep{dong2024rlhf}.
For fair comparison, we also reimplement the Math-Shepherd model with MCTS data generated by Llama-3.1-8B-Instruct, denoted as RLHFlow-8B-LLaMA-Data.
We select the candidate solution with the highest reward-model score as the final output of BoN Sampling.
Moreover, we include a majority-voting baseline, which selects the candidate solution that receives the most votes from the $N$ candidate solutions as the final output.

\subsection{Results}

The experimental results are summarized in Table~\ref{tab:main-exp}.
Our proposed method, \ourrm, consistently outperforms baseline models, including majority voting, across all datasets and generation models.
Notably, \ourrm achieves an average improvement of {6.7\% on MATH-500} and {3.9\% on Olympiad Bench} compared to the strongest baseline model (excluding majority voting).
Interestingly, majority voting outperforms the baseline reward model on MATH-500, suggesting that existing reward models may lack robustness in scoring candidate solutions.
These findings align with previous research~\citep{liu2024rm, kim2024evaluating}, which highlights the limitations of baseline reward models in reliably assessing solution correctness.

\subsection{Difficulty Analysis}
To further investigate scenarios in which the \ourrm outperforms the baseline reward model, we analyze the performance of the \ourrm and the baseline reward model on math problems with different levels of difficulty.
We define the difficulty of a math problem as the fraction of incorrect answers among the candidate solutions:
\begin{equation}
    \text{Difficulty} = \frac{\# \text{incorrect answers}}{\# \text{candidate solutions}}.
\end{equation}
Specifically, we calculate this difficulty when the number of candidate solutions is $n=64$.
We then divide the math problems into four percentile groups based on their difficulty level and evaluate the performance of the \ourrm and baseline models on each percentile in the MATH-500 dataset.
Figure~\ref{fig:difficulty-exp} shows the results.
Except for the easiest problems, the \ourrm consistently outperforms the baseline models across all difficulty levels.
On the challenging problems (Difficulty $> 0.5$), the \ourrm achieves a relative improvement of 40\% to 60\% over the baseline models.
These findings indicate that the \ourrm has strong potential to enhance BoN Sampling on challenging math problems.

\begin{figure}[t]
    \centering
    \includegraphics[width=\linewidth]{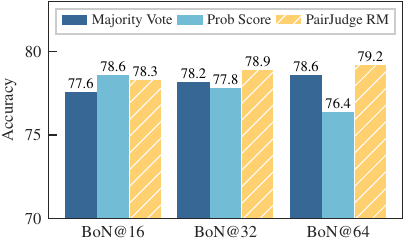}
    \caption{Comparison of the Critic Model with Majority vote/ Prob Score bewteen \ourrm on BoN sampling.
    Accuracy is reported in percentage.}
    \label{fig:bon-critic}
    \vspace*{-0.3cm}
\end{figure}

\begin{figure*}[t]

    \footnotesize
    \centering
    \scalebox{0.9}{
        \begin{tabular}{|p{7cm}|p{7cm}|}
        \toprule
        
        \multicolumn{2}{|p{14.4cm}|}{
        {\textbf{Problem}: A regular hexagon can be divided into six equilateral triangles. If the perimeter of one of the triangles is 21 inches, what is the perimeter, in inches, of the regular hexagon?}
        } \\
        \midrule
        \multicolumn{2}{|p{14.4cm}|}{
        {\textbf{Solution 1}: Since the perimeter of the triangle is 21, each side of the triangle has length \( \frac{21}{3} = 7 \). The hexagon is made up of six of these triangles, so the perimeter of the hexagon is \( 6 \cdot 7 = 42 \). So the final answer is \( \boxed{42} \).
        }} \\
        \midrule
        \textbf{Critic Model} Probabilistic Score: {\color{red}0.0006} & \textbf{\ourrm} Correctness Judgment: {\color{blue}True} \\
        \midrule
        \multicolumn{2}{|p{14.4cm}|}{
        {\textbf{Solution 2}: Since the perimeter of one of the triangles is 21 inches, the side length of the triangle is \( \frac{21}{3} = 7 \) inches. Since there are six equilateral triangles, the perimeter of the hexagon is six times the side length of one triangle, which is \( 6 \cdot 7 = 42 \) inches. So the final answer is \( \boxed{42} \).
        }} \\
        \midrule
        \textbf{Critic Model} Probabilistic Score: {\color{red}0.9999} & \textbf{\ourrm} Correctness Judgment: {\color{blue}True} \\
        \bottomrule
        \end{tabular}
    }
    \vspace{-0.1cm}
    \caption{
    Comparison of the \criticmodel with Probabilistic Score and \ourrm in evaluating two correct solutions to a math problem. 
    The Critic Model assigns drastically different probabilistic scores (0.0006 vs. 0.9999), highlighting its inconsistency, while \ourrm consistently identifies both as correct.
    }
    \label{fig:genrm-vs-pairwise}
    \vspace{-0.1cm}
\end{figure*}

\section{Comparison with \criticmodel}
\label{sec:cov-critic}
\criticmodel~\citep{gao2024critics,mcaleese2024critics}, also known as LLM-as-a-Judge~\citep{zheng2023judging,bai2023benchmarking}, uses one LLM to critique the response of another LLM to a given prompt.
In contrast to ordinary RMs, the output of the \criticmodel is a critique in the form of chain-of-thought reasoning rather than a numerical score.
This setting is similar to that of the \ourrm, as both methods generate a chain-of-thought reasoning text to evaluate the response.
The key difference is that \ourrm performs pairwise judgment, while \criticmodel uses pointwise judgment.
Recently, \criticmodel has been applied in the math and code reasoning domains to verify candidate solutions and assign numerical scores~\citep{mcaleese2024critics,gao2024critics}.
In this section, we compare the effectiveness of \ourrm and \criticmodel in correctness verification and Best-of-N Sampling at the test time.

\subsection{Comparison on Correctness Verification}  
\label{sec:pairwise-comparison}  
We compare the performance of the \ourrm and the \criticmodel on correctness verification.  
Specifically, given one question and two candidate solutions, both models are tasked with judging the correctness of these solutions.  
For a fair comparison, we train the \ourrm and the \criticmodel using the same computational budget and training data.  
In particular, we use the same questions from the MATH-500 training set~\citep{hendrycksmath2021} and candidate solutions generated by Llama-3.1-8B-Instruct~\citep{llama3modelcard} for both models.  
Since each training example for the \ourrm contains two candidate solutions, the training data for the \criticmodel is twice as large.  
All other training details follow Section~\ref{sec:exp-setup}.  

After training, we evaluate both models on the MATH-500 and Olympiad datasets.  
We sample 8,000 candidate solutions from each dataset to form the test set for the \criticmodel.  
To avoid bias, these candidate solutions are generated by Qwen-2.5-7B-Instruct~\citep{yang2024qwen2} using the test split.  
We then pair each solution with another solution that produces a different answer for the same question, yielding 4,000 pairs for evaluating the \ourrm.  

\begin{table}[t]
    \centering
    
    \begin{tabular}{lcccc}
        \toprule
        \bf Model & \bf MATH & \bf Olympiad & \bf Avg. \\
        \midrule
        \criticmodel & 67.7 & 56.9 & 62.3\\
        \ourrm & 70.4 & 64.2 &  67.3\\
        \bottomrule
    \end{tabular}
    \caption{
        Comparison of the \ourrm and LLM-as-a-Judge on the MATH-500 and Olympiad datasets on correctness verification task.
        Candidates are generated by Qwen-2.5-7B-Instruct.
        Accuracy is reported.
        }
        \label{tab:pairwise-comparison}
    \vspace*{-0.3cm}
\end{table}  
As shown in Table~\ref{tab:pairwise-comparison}, the \ourrm outperforms the \criticmodel on both the MATH-500 and Olympiad datasets.  
This result suggests that pairwise judgment is more effective than pointwise judgment when judging correctness.  
Notably, on the more challenging Olympiad dataset, the \ourrm achieves a larger improvement, highlighting its potential for difficult math problems.  

\subsection{Comparison on Best-of-N Sampling}  
\label{sec:llm-judge-exp}  
As described in Section~\ref{sec:pairwise-comparison}, the primary difference between the \ourrm and the \criticmodel is the judgment process.  
The \ourrm performs pairwise judgment, while the \criticmodel performs pointwise judgment.  
This difference makes it challenging to apply the \criticmodel for BoN Sampling at test time.  
If the \criticmodel verifies two candidate solutions with different answers as correct, it remains unclear which one is superior.  
In contrast, the \ourrm directly compares correctness to select the better solution.  

To enable BoN Sampling with the \criticmodel, two approaches are typically adopted.  
\paragraph{Combine with Majority Voting}  
One option is to combine the \criticmodel with majority voting.  
The \criticmodel first judges the correctness of each candidate and removes those marked as incorrect.
Majority voting is then applied to the remaining solutions to determine the final output. 

\paragraph{Use Probabilistic Score}  
Another approach is to use the probabilistic score assigned by the \criticmodel to each candidate solution.  
In this method, the \criticmodel is prompted to generate a token—either “correct” or “incorrect”—within its reasoning text to indicate correctness.  
\citet{zhang2024generative} suggest that the probability of generating the token “correct” can serve as the score for each candidate solution.  
The candidate with the highest score is then selected as the final output.  

To prevent data leakage, we use the MATH-500 test split and candidate solutions generated by Qwen-2.5-7B-Instruct~\citep{yang2024qwen2} to evaluate both models on BoN Sampling.  
For a fair comparison, we reuse the \criticmodel and \ourrm trained in Section~\ref{sec:pairwise-comparison} for this evaluation.  

As shown in Figure~\ref{fig:bon-critic}, the \ourrm consistently outperforms the \criticmodel on the MATH-500 dataset.  
This result demonstrates that, under the same training budget, the \ourrm is more effective at BoN Sampling than the \criticmodel.  
Additionally, the \criticmodel using a probabilistic score underperforms compared to majority voting, likely due to its tendency to assign highly polarized scores.  
As illustrated in Figure~\ref{fig:genrm-vs-pairwise}, even similar candidate solutions receive substantially different probabilistic scores.  
This observation suggests that the probabilistic scoring method faces robustness and stability issues similar to those observed in reward models~\citep{liu2024rm,kim2024evaluating}.

\section{Related Work}

\subsection{Test Time Scaling and Best-of-N Sampling}
Test-time scaling, introduced with \texttt{o1}\citep{openai_o1_preview}, improves model performance during inference by dedicating more computation~\citep{snell2024scaling, wu2024empirical}
Approaches include Monte Carlo Tree Search~\citep{zhang2024accessing, gao2024interpretable} and long-chain-of-thought~\citep{min2024imitate, qwq}
Best-of-N Sampling is one such approach, generating $N$ candidate solutions and selecting the best via a reward model~\citep{wang2023math, lightman2023let, wang2024openr, zhang2024generative}
Performing BoN sampling with tournament-style selection was first introduced for instruction-following tasks~\citep{lee-etal-2024-aligning}
Our knockout method differs by grouping candidates with identical answers to avoid unnecessary comparisons and using generative judgment via CoT reasoning~\citep{wei2022chain} rather than discriminative scoring.

\subsection{Reward Models and Critic Models}
Reward models (RMs) assign numerical scores to LLM outputs for feedback during training and inference~\citep{lambert2024rewardbench, liu2024rm, wang2024secrets, lightman2023let}.  
They can operate in pointwise or pairwise manners~\citep{yuan2024advancing, llm-blender-2023}.  
Critic models evaluate response quality, especially in reasoning tasks like math and code~\citep{gao2024critics, mcaleese2024critics, zheng2023judging, li2024llmasajudge}.  
Unlike RMs, Critic Models function offering textual feedback instead of numerical score.
\ourrm differs from them by simultaneously judging the correctness of two responses using chain-of-thought reasoning instead of pointwise evaluation like Critic Models and numerical scoring like RMs.

\section{Conclusion}
We propose the Pairwise Judge Reward Model (\ourrm) with a knockout tournament for BoN Sampling. 
\ourrm uses CoT reasoning to simultaneously evaluate two candidate solutions, eliminating arbitrary scoring and enabling cross-validation. 
The knockout tournament iteratively eliminates incorrect solutions through pairwise judgments. 
We create a 432K pairwise judgment dataset to train \ourrm. 
Experiments show \ourrm significantly outperforms baseline RMs in BoN Sampling on benchmarks.

\section*{Limitation}
\label{sec:limitation}
The main limitation of the proposed method lies on the inference time.
To conduct the BoN Sampling with the \ourrm, serval rounds of pairwise verification are required to select the best candidate solution.
This process is time-consuming and may not be suitable for latency-sensitive applications.
However, the proposed method can be potentially accelerated by parallel computing or other optimization techniques to reduce the inference time.
For example, the multiple pairwise verification can easily be parallelized to multiple GPUs to speed up the inference process since they are independent of each other.
Moreover, with popularization of the inference-time scaling, it is a common practice to increase the computational resources to improve the performance of the model in solving complex reasoning tasks like math problems~\citep{snell2024scaling, wu2024empirical}.

\section*{Ethical Considerations}
\label{sec:ethical-considerations}
In this work, all the data and models are acquired from public datasets and pre-trained models, and no human subjects are involved in the experiments. 
Considering the potential hallucination and bias in the pre-trained models, it is worth nothing that the user should be cautious when applying the proposed method to real-world applications such as use \ourrm to check human student's homework in the educational system.

\bibliography{custom}

\begin{thebibliography}{43}
\providecommand{\natexlab}[1]{#1}

\bibitem[{Abdul~Gafoor and Jisha(2014)}]{abdul2014study}
K~Abdul~Gafoor and P~Jisha. 2014.
\newblock A study of reliability of marking and absolute grading in secondary
  schools.
\newblock \emph{Online Submission}, 2(2):292--298.

\bibitem[{AI(2025)}]{meta2025llama31}
Meta AI. 2025.
\newblock \href {https://ai.meta.com/blog/meta-llama-3-1/} {Introducing llama
  3.1: Our most capable models to date}.
\newblock Accessed: 2025-02-01.

\bibitem[{AI@Meta(2024)}]{llama3modelcard}
AI@Meta. 2024.
\newblock \href {https://github.com/meta-llama/llama3/blob/main/MODEL_CARD.md}
  {Llama 3 model card}.

\bibitem[{Bai et~al.(2023)Bai, Ying, Cao, Lv, He, Wang, Yu, Zeng, Xiao, Lyu,
  Zhang, Li, and Hou}]{bai2023benchmarking}
Yushi Bai, Jiahao Ying, Yixin Cao, Xin Lv, Yuze He, Xiaozhi Wang, Jifan Yu,
  Kaisheng Zeng, Yijia Xiao, Haozhe Lyu, Jiayin Zhang, Juanzi Li, and Lei Hou.
  2023.
\newblock \href {https://openreview.net/forum?id=IiRHQ7gvnq} {Benchmarking
  foundation models with language-model-as-an-examiner}.
\newblock In \emph{Thirty-seventh Conference on Neural Information Processing
  Systems Datasets and Benchmarks Track}.

\bibitem[{Beygelzimer et~al.(2021)Beygelzimer, Dauphin, Liang, and
  Vaughan}]{neurips2021}
Alina Beygelzimer, Yann Dauphin, Percy Liang, and Jennifer~Wortman Vaughan.
  2021.
\newblock The neurips 2021 consistency experiment.
\newblock
  \url{https://blog.neurips.cc/2021/12/08/the-neurips-2021-consistency-experiment/}.
\newblock Accessed: 2024-12-26.

\bibitem[{Devriesere et~al.(2024)Devriesere, Csató, and
  Goossens}]{Devriesere2024}
Karel Devriesere, László Csató, and Dries Goossens. 2024.
\newblock \href {https://doi.org/10.1016/j.ejor.2024.10.053} {Tournament
  design: A review from an operational research perspective}.
\newblock \emph{European Journal of Operational Research}, In Press, Corrected
  Proof.
\newblock Available online 9 November 2024.

\bibitem[{Dong et~al.(2024)Dong, Xiong, Pang, Wang, Zhao, Zhou, Jiang, Sahoo,
  Xiong, and Zhang}]{dong2024rlhf}
Hanze Dong, Wei Xiong, Bo~Pang, Haoxiang Wang, Han Zhao, Yingbo Zhou, Nan
  Jiang, Doyen Sahoo, Caiming Xiong, and Tong Zhang. 2024.
\newblock \href {https://openreview.net/forum?id=a13aYUU9eU} {{RLHF} workflow:
  From reward modeling to online {RLHF}}.
\newblock \emph{Transactions on Machine Learning Research}.

\bibitem[{Gao et~al.(2024{\natexlab{a}})Gao, Cai, Xu, Wang, Zheng, Lin, Lu,
  Lin, Zhou, Xiao, Hu, Liu, and Chang}]{gao2024critics}
Bofei Gao, Zefan Cai, Runxin Xu, Peiyi Wang, Ce~Zheng, Runji Lin, Keming Lu,
  Junyang Lin, Chang Zhou, Wen Xiao, Junjie Hu, Tianyu Liu, and Baobao Chang.
  2024{\natexlab{a}}.
\newblock \href {https://doi.org/10.48550/arXiv.2406.14024} {Llm critics help
  catch bugs in mathematics: Towards a better mathematical verifier with
  natural language feedback}.
\newblock \emph{CoRR}, abs/2406.14024.

\bibitem[{Gao et~al.(2024{\natexlab{b}})Gao, Niu, He, Xu, Liu, Liu, Hu, and
  Wen}]{gao2024interpretable}
Zitian Gao, Boye Niu, Xuzheng He, Haotian Xu, Hongzhang Liu, Aiwei Liu, Xuming
  Hu, and Lijie Wen. 2024{\natexlab{b}}.
\newblock \href {https://arxiv.org/abs/2410.01707} {Interpretable contrastive
  monte carlo tree search reasoning}.
\newblock \emph{Preprint}, arXiv:2410.01707.

\bibitem[{He et~al.(2024)He, Luo, Bai, Hu, Thai, Shen, Hu, Han, Huang, Zhang,
  Liu, Qi, Liu, and Sun}]{he-etal-2024-olympiadbench}
Chaoqun He, Renjie Luo, Yuzhuo Bai, Shengding Hu, Zhen Thai, Junhao Shen, Jinyi
  Hu, Xu~Han, Yujie Huang, Yuxiang Zhang, Jie Liu, Lei Qi, Zhiyuan Liu, and
  Maosong Sun. 2024.
\newblock \href {https://doi.org/10.18653/v1/2024.acl-long.211}
  {{O}lympiad{B}ench: A challenging benchmark for promoting {AGI} with
  olympiad-level bilingual multimodal scientific problems}.
\newblock In \emph{Proceedings of the 62nd Annual Meeting of the Association
  for Computational Linguistics (Volume 1: Long Papers)}, pages 3828--3850,
  Bangkok, Thailand. Association for Computational Linguistics.

\bibitem[{Hendrycks et~al.(2021)Hendrycks, Burns, Kadavath, Arora, Basart,
  Tang, Song, and Steinhardt}]{hendrycksmath2021}
Dan Hendrycks, Collin Burns, Saurav Kadavath, Akul Arora, Steven Basart, Eric
  Tang, Dawn Song, and Jacob Steinhardt. 2021.
\newblock {Measuring Mathematical Problem Solving With the MATH Dataset}.
\newblock \emph{NeurIPS}.

\bibitem[{Hoffmann et~al.(2022)Hoffmann, Borgeaud, Mensch, Buchatskaya, Cai,
  Rutherford, de~las Casas, Hendricks, Welbl, Clark, Hennigan, Noland,
  Millican, van~den Driessche, Damoc, Guy, Osindero, Simonyan, Elsen, Vinyals,
  Rae, and Sifre}]{hoffmann2022an}
Jordan Hoffmann, Sebastian Borgeaud, Arthur Mensch, Elena Buchatskaya, Trevor
  Cai, Eliza Rutherford, Diego de~las Casas, Lisa~Anne Hendricks, Johannes
  Welbl, Aidan Clark, Tom Hennigan, Eric Noland, Katherine Millican, George
  van~den Driessche, Bogdan Damoc, Aurelia Guy, Simon Osindero, Karen Simonyan,
  Erich Elsen, Oriol Vinyals, Jack~William Rae, and Laurent Sifre. 2022.
\newblock \href {https://openreview.net/forum?id=iBBcRUlOAPR} {An empirical
  analysis of compute-optimal large language model training}.
\newblock In \emph{Advances in Neural Information Processing Systems}.

\bibitem[{Jiang et~al.(2023)Jiang, Ren, and Lin}]{llm-blender-2023}
Dongfu Jiang, Xiang Ren, and Bill~Yuchen Lin. 2023.
\newblock Llm-blender: Ensembling large language models with pairwise
  comparison and generative fusion.
\newblock In \emph{Proceedings of the 61th Annual Meeting of the Association
  for Computational Linguistics (ACL 2023)}.

\bibitem[{Jonsson and Svingby(2007)}]{jonsson2007use}
Anders Jonsson and Gunilla Svingby. 2007.
\newblock The use of scoring rubrics: Reliability, validity and educational
  consequences.
\newblock \emph{Educational research review}, 2(2):130--144.

\bibitem[{Kim et~al.(2024)Kim, Kang, Kwon, Chae, Won, Lee, and
  Yeo}]{kim2024evaluating}
Sunghwan Kim, Dongjin Kang, Taeyoon Kwon, Hyungjoo Chae, Jungsoo Won, Dongha
  Lee, and Jinyoung Yeo. 2024.
\newblock Evaluating robustness of reward models for mathematical reasoning.
\newblock \emph{arXiv preprint arXiv:2410.01729}.

\bibitem[{Lambert et~al.(2024)Lambert, Pyatkin, Morrison, Miranda, Lin, Chandu,
  Dziri, Kumar, Zick, Choi et~al.}]{lambert2024rewardbench}
Nathan Lambert, Valentina Pyatkin, Jacob Morrison, LJ~Miranda, Bill~Yuchen Lin,
  Khyathi Chandu, Nouha Dziri, Sachin Kumar, Tom Zick, Yejin Choi, et~al. 2024.
\newblock Rewardbench: Evaluating reward models for language modeling.
\newblock \emph{arXiv preprint arXiv:2403.13787}.

\bibitem[{Lee et~al.(2024)Lee, Kim, Yousefpour, Seo, Yoo, and
  Yu}]{lee-etal-2024-aligning}
Sangkyu Lee, Sungdong Kim, Ashkan Yousefpour, Minjoon Seo, Kang~Min Yoo, and
  Youngjae Yu. 2024.
\newblock \href {https://doi.org/10.18653/v1/2024.acl-long.617} {Aligning large
  language models by on-policy self-judgment}.
\newblock In \emph{Proceedings of the 62nd Annual Meeting of the Association
  for Computational Linguistics (Volume 1: Long Papers)}, pages 11442--11459,
  Bangkok, Thailand. Association for Computational Linguistics.

\bibitem[{Li et~al.(2024)Li, Jiang, Huang, Beigi, Zhao, Tan, Bhattacharjee,
  Jiang, Chen, Wu, Shu, Cheng, and Liu}]{li2024llmasajudge}
Dawei Li, Bohan Jiang, Liangjie Huang, Alimohammad Beigi, Chengshuai Zhao, Zhen
  Tan, Amrita Bhattacharjee, Yuxuan Jiang, Canyu Chen, Tianhao Wu, Kai Shu,
  Lu~Cheng, and Huan Liu. 2024.
\newblock From generation to judgment: Opportunities and challenges of
  llm-as-a-judge.
\newblock \emph{arXiv preprint arXiv: 2411.16594}.

\bibitem[{LI et~al.(2024)LI, Beeching, Tunstall, Lipkin, Soletskyi, Huang,
  Rasul, Yu, Jiang, Shen, Qin, Dong, Zhou, Fleureau, Lample, and
  Polu}]{numina_math_datasets}
Jia LI, Edward Beeching, Lewis Tunstall, Ben Lipkin, Roman Soletskyi,
  Shengyi~Costa Huang, Kashif Rasul, Longhui Yu, Albert Jiang, Ziju Shen, Zihan
  Qin, Bin Dong, Li~Zhou, Yann Fleureau, Guillaume Lample, and Stanislas Polu.
  2024.
\newblock Numinamath.
\newblock
  \url{[https://huggingface.co/AI-MO/NuminaMath-CoT](https://github.com/project-numina/aimo-progress-prize/blob/main/report/numina_dataset.pdf)}.

\bibitem[{Lightman et~al.(2023)Lightman, Kosaraju, Burda, Edwards, Baker, Lee,
  Leike, Schulman, Sutskever, and Cobbe}]{lightman2023let}
Hunter Lightman, Vineet Kosaraju, Yura Burda, Harri Edwards, Bowen Baker, Teddy
  Lee, Jan Leike, John Schulman, Ilya Sutskever, and Karl Cobbe. 2023.
\newblock {Let's Verify Step by Step}.
\newblock \emph{arXiv preprint arXiv:2305.20050}.

\bibitem[{Liu and Zeng(2024)}]{skyworkreward2024}
Chris~Yuhao Liu and Liang Zeng. 2024.
\newblock \href {https://huggingface.co/Skywork} {Skywork reward model series}.
\newblock \url{https://huggingface.co/Skywork}.

\bibitem[{Liu et~al.(2024)Liu, Yao, Min, Cao, Hou, and Li}]{liu2024rm}
Yantao Liu, Zijun Yao, Rui Min, Yixin Cao, Lei Hou, and Juanzi Li. 2024.
\newblock Rm-bench: Benchmarking reward models of language models with subtlety
  and style.
\newblock \emph{arXiv preprint arXiv:2410.16184}.

\bibitem[{McAleese et~al.(2024)McAleese, Pokorny, Uribe, Nitishinskaya,
  Trebacz, and Leike}]{mcaleese2024critics}
Nat McAleese, Rai~Michael Pokorny, Juan Felipe~Ceron Uribe, Evgenia
  Nitishinskaya, Maja Trebacz, and Jan Leike. 2024.
\newblock Llm critics help catch llm bugs.
\newblock \emph{arXiv preprint arXiv:2407.00215}.

\bibitem[{Min et~al.(2024)Min, Chen, Jiang, Chen, Deng, Hu, Tang, Wang, Cheng,
  Song et~al.}]{min2024imitate}
Yingqian Min, Zhipeng Chen, Jinhao Jiang, Jie Chen, Jia Deng, Yiwen Hu, Yiru
  Tang, Jiapeng Wang, Xiaoxue Cheng, Huatong Song, et~al. 2024.
\newblock Imitate, explore, and self-improve: A reproduction report on
  slow-thinking reasoning systems.
\newblock \emph{arXiv preprint arXiv:2412.09413}.

\bibitem[{OpenAI(2024)}]{openai_o1_preview}
OpenAI. 2024.
\newblock Introducing openai o1 preview.
\newblock \url{https://openai.com/index/introducing-openai-o1-preview/}.
\newblock Accessed: 2024-09-17.

\bibitem[{Paszke et~al.(2019)Paszke, Gross, Massa, Lerer, Bradbury, Chanan,
  Killeen, Lin, Gimelshein, Antiga, Desmaison, Kopf, Yang, DeVito, Raison,
  Tejani, Chilamkurthy, Steiner, Fang, Bai, and Chintala}]{paszke2019pytorch}
Adam Paszke, Sam Gross, Francisco Massa, Adam Lerer, James Bradbury, Gregory
  Chanan, Trevor Killeen, Zeming Lin, Natalia Gimelshein, Luca Antiga, Alban
  Desmaison, Andreas Kopf, Edward Yang, Zachary DeVito, Martin Raison, Alykhan
  Tejani, Sasank Chilamkurthy, Benoit Steiner, Lu~Fang, Junjie Bai, and Soumith
  Chintala. 2019.
\newblock \href
  {https://papers.nips.cc/paper/9015-pytorch-an-imperative-style-high-performance-deep-learning-library.pdf}
  {Pytorch: An imperative style, high-performance deep learning library}.
\newblock In \emph{Advances in Neural Information Processing Systems 32}, pages
  8024--8035. Curran Associates, Inc.
\newblock Accessed: 2025-02-12.

\bibitem[{Qwen~Team(2024)}]{yang2024qwen2}
Alibaba Qwen~Team. 2024.
\newblock Qwen2. 5 technical report.
\newblock \emph{arXiv preprint arXiv:2412.15115}.

\bibitem[{{Qwen Team, Alibaba}(2023)}]{qwq}
{Qwen Team, Alibaba}. 2023.
\newblock {QwQ-32B Preview}.
\newblock \url{https://qwenlm.github.io/zh/blog/qwq-32b-preview/}.
\newblock Accessed: 2024-12-28.

\bibitem[{Rasley et~al.(2020)Rasley, He, Song, and
  Smelyanskiy}]{rasley2020deepspeed}
Jeffrey Rasley, Zhizhou He, Jianwen Song, and Mikhail Smelyanskiy. 2020.
\newblock Deepspeed: Extreme-scale model training for everyone.
\newblock \url{https://www.deepspeed.ai}.
\newblock Accessed: 2025-02-12.

\bibitem[{Snell et~al.(2024)Snell, Lee, Xu, and Kumar}]{snell2024scaling}
Charlie Snell, Jaehoon Lee, Kelvin Xu, and Aviral Kumar. 2024.
\newblock Scaling llm test-time compute optimally can be more effective than
  scaling model parameters.
\newblock \emph{arXiv preprint arXiv:2408.03314}.

\bibitem[{Wang et~al.(2024{\natexlab{a}})Wang, Zheng, Chen, Liu, Dou, Huang,
  Shen, Jin, Zhou, Shi, Gao, Xu, Zhou, Fan, Xi, Zhao, Wang, Ji, Yan, Shen,
  Chen, Gui, Zhang, Qiu, Huang, Wu, and Jiang}]{wang2024secrets}
Binghai Wang, Rui Zheng, Lu~Chen, Yan Liu, Shihan Dou, Caishuang Huang, Wei
  Shen, Senjie Jin, Enyu Zhou, Chenyu Shi, Songyang Gao, Nuo Xu, Yuhao Zhou,
  Xiaoran Fan, Zhiheng Xi, Jun Zhao, Xiao Wang, Tao Ji, Hang Yan, Lixing Shen,
  Zhan Chen, Tao Gui, Qi~Zhang, Xipeng Qiu, Xuanjing Huang, Zuxuan Wu, and
  Yu-Gang Jiang. 2024{\natexlab{a}}.
\newblock \href {https://arxiv.org/abs/2401.06080} {Secrets of rlhf in large
  language models part ii: Reward modeling}.
\newblock \emph{Preprint}, arXiv:2401.06080.

\bibitem[{Wang et~al.(2024{\natexlab{b}})Wang, Fang, Wan, Wen, Zhu, Liu, Gong,
  Song, Chen, Ni et~al.}]{wang2024openr}
Jun Wang, Meng Fang, Ziyu Wan, Muning Wen, Jiachen Zhu, Anjie Liu, Ziqin Gong,
  Yan Song, Lei Chen, Lionel~M Ni, et~al. 2024{\natexlab{b}}.
\newblock Openr: An open source framework for advanced reasoning with large
  language models.
\newblock \emph{arXiv preprint arXiv:2410.09671}.

\bibitem[{Wang et~al.(2023)Wang, Li, Shao, Xu, Dai, Li, Chen, Wu, and
  Sui}]{wang2023math}
Peiyi Wang, Lei Li, Zhihong Shao, RX~Xu, Damai Dai, Yifei Li, Deli Chen, Y~Wu,
  and Zhifang Sui. 2023.
\newblock Math-shepherd: A label-free step-by-step verifier for llms in
  mathematical reasoning.
\newblock \emph{arXiv preprint arXiv:2312.08935}.

\bibitem[{Wei et~al.(2022)Wei, Wang, Schuurmans, Bosma, brian ichter, Xia, Chi,
  Le, and Zhou}]{wei2022chain}
Jason Wei, Xuezhi Wang, Dale Schuurmans, Maarten Bosma, brian ichter, Fei Xia,
  Ed~H. Chi, Quoc~V Le, and Denny Zhou. 2022.
\newblock \href {https://openreview.net/forum?id=_VjQlMeSB_J} {Chain of thought
  prompting elicits reasoning in large language models}.
\newblock In \emph{Advances in Neural Information Processing Systems}.

\bibitem[{Wolf et~al.(2020)Wolf, Debut, Sanh, Chaumond, Delangue, and
  Maisonneuve}]{wolf2020transformers}
Thomas Wolf, Lysandre Debut, Victor Sanh, Julien Chaumond, Clément Delangue,
  and Alexander Maisonneuve. 2020.
\newblock Hugging face's transformers: State-of-the-art natural language
  processing.
\newblock \url{https://huggingface.co/transformers}.
\newblock Accessed: 2025-02-12.

\bibitem[{Wu et~al.(2024)Wu, Sun, Li, Welleck, and Yang}]{wu2024empirical}
Yangzhen Wu, Zhiqing Sun, Shanda Li, Sean Welleck, and Yiming Yang. 2024.
\newblock An empirical analysis of compute-optimal inference for
  problem-solving with language models.
\newblock \emph{arXiv preprint arXiv:2408.00724}.

\bibitem[{Ye et~al.(2024)Ye, Xiong, Zhang, Dong, Jiang, and
  Zhang}]{ye2024online}
Chenlu Ye, Wei Xiong, Yuheng Zhang, Hanze Dong, Nan Jiang, and Tong Zhang.
  2024.
\newblock \href {https://openreview.net/forum?id=TwdX1W3M6S} {Online iterative
  reinforcement learning from human feedback with general preference model}.
\newblock In \emph{The Thirty-eighth Annual Conference on Neural Information
  Processing Systems}.

\bibitem[{Yuan et~al.(2024{\natexlab{a}})Yuan, Cui, Wang, Ding, Wang, Deng,
  Shan, Chen, Xie, Lin, Liu, Zhou, Peng, Liu, and Sun}]{yuan2024advancing}
Lifan Yuan, Ganqu Cui, Hanbin Wang, Ning Ding, Xingyao Wang, Jia Deng, Boji
  Shan, Huimin Chen, Ruobing Xie, Yankai Lin, Zhenghao Liu, Bowen Zhou, Hao
  Peng, Zhiyuan Liu, and Maosong Sun. 2024{\natexlab{a}}.
\newblock \href {https://arxiv.org/abs/2404.02078} {Advancing llm reasoning
  generalists with preference trees}.
\newblock \emph{Preprint}, arXiv:2404.02078.

\bibitem[{Yuan et~al.(2024{\natexlab{b}})Yuan, Li, Chen, Cui, Ding, Zhang,
  Zhou, Liu, and Peng}]{yuan2024free}
Lifan Yuan, Wendi Li, Huayu Chen, Ganqu Cui, Ning Ding, Kaiyan Zhang, Bowen
  Zhou, Zhiyuan Liu, and Hao Peng. 2024{\natexlab{b}}.
\newblock Free process rewards without process labels.
\newblock \emph{arXiv preprint arXiv:2412.01981}.

\bibitem[{Zhang et~al.(2024{\natexlab{a}})Zhang, Huang, Zhou, Li, and
  Ouyang}]{zhang2024accessing}
Di~Zhang, Xiaoshui Huang, Dongzhan Zhou, Yuqiang Li, and Wanli Ouyang.
  2024{\natexlab{a}}.
\newblock Accessing gpt-4 level mathematical olympiad solutions via monte carlo
  tree self-refine with llama-3 8b.
\newblock \emph{arXiv preprint arXiv:2406.07394}.

\bibitem[{Zhang et~al.(2024{\natexlab{b}})Zhang, Hosseini, Bansal, Kazemi,
  Kumar, and Agarwal}]{zhang2024generative}
Lunjun Zhang, Arian Hosseini, Hritik Bansal, Mehran Kazemi, Aviral Kumar, and
  Rishabh Agarwal. 2024{\natexlab{b}}.
\newblock Generative verifiers: Reward modeling as next-token prediction.
\newblock \emph{arXiv preprint arXiv:2408.15240}.

\bibitem[{Zhao et~al.(2023)Zhao, Joshi, Liu, Khalman, Saleh, and
  Liu}]{zhao2023slic}
Yao Zhao, Rishabh Joshi, Tianqi Liu, Misha Khalman, Mohammad Saleh, and Peter~J
  Liu. 2023.
\newblock Slic-hf: Sequence likelihood calibration with human feedback.
\newblock \emph{arXiv preprint arXiv:2305.10425}.

\bibitem[{Zheng et~al.(2023)Zheng, Chiang, Sheng, Zhuang, Wu, Zhuang, Lin, Li,
  Li, Xing et~al.}]{zheng2023judging}
Lianmin Zheng, Wei-Lin Chiang, Ying Sheng, Siyuan Zhuang, Zhanghao Wu, Yonghao
  Zhuang, Zi~Lin, Zhuohan Li, Dacheng Li, Eric Xing, et~al. 2023.
\newblock {Judging LLM-as-a-judge with MT-Bench and Chatbot Arena}.
\newblock \emph{arXiv preprint arXiv:2306.05685}.

\end{thebibliography}

\newpage
\appendix

\section{Dataset Information}

In this section, we provide detailed information about the datasets used in this work. The datasets employed for evaluating mathematical reasoning tasks come from various sources, each contributing unique characteristics for comprehensive benchmarking.

\paragraph{MATH-500} is a collection of problems designed to test mathematical reasoning capabilities. 
It covers a wide range of mathematical concepts and problem types, from basic algebra to more advanced topics. 
Specifically, MATH-500 is newer, IID version of MATH~\citep{hendrycksmath2021}, which is a widely used benchmark for avoide the data leakage issue.
\textbf{Size:} 500 problems.
\textbf{Source:} \url{https://github.com/openai/prm800k/tree/main/prm800k/math_splits}. 
\textbf{License:} MIT License.

\paragraph{Olympiad Bench} is derived from a collection of problems from various international mathematical olympiads. It includes a broad range of challenging problems, covering topics like number theory, combinatorics, geometry, and algebra. This dataset is particularly useful for testing a model's ability to handle competition-level mathematical reasoning tasks.
\textbf{Size:} 8,476 problems
\textbf{Source:} \url{https://arxiv.org/abs/2402.14008}.
\textbf{License:} MIT License

\paragraph{NuminaMath-CoT} is a dataset that includes 860k math problems, where each solution is formatted in a Chain of Thought (CoT) manner. The sources of the dataset range from Chinese high school math exercises to US and international mathematics olympiad competition problems. The data were primarily collected from online exam paper PDFs and mathematics discussion forums. The processing steps include (a) OCR from the original PDFs, (b) segmentation into problem-solution pairs, (c) Translation into English, (d) realignment to produce a CoT reasoning format, and (e) final answer formatting.
\textbf{Size:} 860k problems.
\textbf{Source:} \url{https://huggingface.co/datasets/AI-MO/NuminaMath-CoT}.
\textbf{License:} Apache License 2.0.

\section{Computational Resources}

This section provides detailed information about the computational resources used in our experiments. The training of Qwen-2.5-7B-Instr on the \ourdataset was conducted on an 8-GPU H100 server, with an estimated training duration of approximately 24 hours. 
The construction of the \ourdataset relies on the \texttt{gemini-1.5-flash} API server through Google Cloud.
The associated API costs for this project amounted to approximately 2,000 USD.

\section{Future Work}
\subsection*{Application in Reinforcement Learning}
In this work, we mainly focus on how to perform the BoN Sampling at test time with the \ourrm.
This experiment setting follows the existing work~\citep{wang2023math, lightman2023let, wang2024openr, zhang2024generative} and helps us to compare the performance of the \ourrm with baseline models and verify the effectiveness of the proposed method.
However, the \ourrm can also be applied at the Reinforcement Learning (RL) training stage to improve the performance of the model in solving complex reasoning tasks like math problems.
To apply in the training stage, the \ourrm need to assign a numerical score to the candidate solutions just like the discriminative reward model~\citep{lambert2024rewardbench,liu2024rm}.
Such a numerical score could acquired by the winning rate of the candidate solutions in the knockout tournament, which can be used as the reward signal to guide the training of the model.
In the future, we plan to explore the application of the \ourrm in the RL training stage to improve the performance of the model in solving complex reasoning tasks like math problems.

\subsection*{Alternative Tournament Strategies}
In this work, we introduce the knockout tournament to select the best candidate solution, where the candidate solutions are viewed as players in the tournament and each pairwise comparison is viewed as a match between two players.
The main reason for choosing the knockout tournament is that it is one of the most naive tournament design that could select the best candidate under time complexity $O(N)$, where $N$ is the number of candidate solutions.
It worth noting that there are tons of alternative tournament strategies that could be used to select the best candidate solution, such as the round-robin tournament, the Swiss-system tournament, and the double-elimination tournament~\citep{Devriesere2024}
Such alternative tournament strategies could be potentially used to improve the performance of the \ourrm in selecting the best candidate solution, and we plan to explore the application of the alternative tournament strategies in the future work.

\section{Potential Improvement}
Due to the computational limitation and resource constraints, there are several potential improvements that could be made to further improve the performance of the \ourrm.
\begin{itemize}[noitemsep,topsep=0pt,parsep=0pt,partopsep=0pt]
    \item \textbf{Bigger Model Capacity:} Due to the computational limitation, 
    although we presents a promising and scalable dataset contruction in Section~\ref{sec:dataset}, the \ourrm is trained with the Qwen-2.5-7B-Instr, which is a relatively small model compared to the state-of-the-art models like the Qwen-2.5-70B-Instr, LLaMA-3.1-70B-Instr and QwQ-32B.
    According to the Chinchilla Law~\citep{hoffmann2022an}, under the same training data and training time, a model with larger capacity can achieve better performance than a model with smaller capacity.
    \item \textbf{More Data Scaling Dimension:} The \ourrm is trained with the \ourdataset dataset, which contains 343K training data for the \ourrm. Now there are two directions to further improve the performance of the \ourrm: 1) use more models rather than only LLama-3.1-8B-Instruct to generate the candidate solutions, and 2) Use more models rather than only \texttt{gemini-1.5-flash} to annotate the training data. This two directions could potentially magnitudes the size of the training data and improve the performance of the \ourrm.
    \item \textbf{Long-Cot Base Model:} The \ourrm is trained with the Qwen-2.5-7B-Instruct. Considering the recent success of the Long-Cot models such as the QwQ-32B~\citep{qwq} in reasoning task, it is worth exploring the application of the \ourrm with the Long-Cot models to further improve the performance of the \ourrm.

\end{itemize}

\section{Artifacts in Our Research}

Our work is built upon several key artifacts that have played a crucial role in enabling the development and evaluation of the proposed Pairwise Judge Reward Model (\ourrm). 

\textbf{PyTorch} \cite{paszke2019pytorch} is a widely used deep learning framework that provides flexible and efficient tools for building neural networks, making it an essential artifact for modern NLP research. Its dynamic computational graph and GPU acceleration have been pivotal in enabling rapid prototyping and experimentation, especially in the context of transformer-based models and reinforcement learning.

The \textbf{Hugging Face Transformers} library \cite{wolf2020transformers} is another critical tool that has revolutionized NLP. It provides an extensive collection of pre-trained models and tools for working with transformer architectures, making it easier for researchers and practitioners to fine-tune models on domain-specific tasks. The library's user-friendly API, combined with its comprehensive model hub, has democratized access to state-of-the-art models such as BERT, GPT, T5, and more.

\textbf{DeepSpeed} \cite{rasley2020deepspeed} is a library developed by Microsoft that aims to optimize large-scale model training. It introduces techniques such as mixed-precision training and model parallelism to significantly reduce memory usage and speed up training, making it an essential tool for training large transformer models. DeepSpeed's support for efficient distributed training has enabled researchers to scale up their models while reducing computational costs.

The \textbf{Llama} model \cite{llama3modelcard} is a family of large language models developed by Meta AI. Trained on approximately 15 trillion tokens, Llama models are available in sizes ranging from 8 billion to 405 billion parameters. They have demonstrated superior performance across various NLP benchmarks, making them a valuable resource for tasks such as text generation, translation, and summarization.

The \textbf{Qwen} model \cite{yang2024qwen2} is a series of large language models developed by Alibaba Cloud. The Qwen series includes models with varying parameter counts ranging from 0.5B to 72B. These models have shown competitive performance across diverse benchmarks, including language understanding, generation, multilingual proficiency, coding, mathematics, and reasoning. The Qwen series has been instrumental in advancing the capabilities of LLMs in various applications.

All these artifacts have played a crucial role in enabling our research and have significantly contributed to our research.

\section{Experiments Replication}
To facilitate the replication of our experiments, we plan to release the codebase, model checkpoints, and dataset used in this work. 
The codebase will be made available on GitHub, along with detailed instructions on how to reproduce the experiments.
As for the model checkpoints and dataset, we plan to provide in Huggingface.
All the experiments will be conducted on three times and the average results will be reported to minimize the randomness in the experiments.

\section{Prompt Templates}
Here we provide the prompt templates for the \ourrm in Table~\ref{tab:pairwise-prompt}
, which is used to guide the \ourrm to judge the correctness of two candidate solutions to a given math problem.

\section{Math Problem Filtering Criteria}
Here we provide the filtering criteria applied to the dataset in Table~\ref{tab:math-problem-filter}, which is used to remove low-quality, proof-based, or multiple-choice problems.
\newpage
\begin{table*}

    \centering
    \caption{
        \small
        Prompt Template for \ourrm, the \green{\{question\}}, \green{\{response\_a\}}, and \green{\{response\_b\}} are placeholders for the math question, response A, and response B, respectively.
    }
    \label{tab:pairwise-prompt}
    \fontsize{8}{9}\selectfont 
    \begin{tabular}{p{\linewidth}}
        \toprule

        \textbf{Task Objective:} \\
        Evaluate the correctness of two responses (Response A and Response B) to a given math question. Perform a step-by-step verification of each response’s accuracy. After completing the step-by-step checks, provide a final correctness judgment for each response. \\
        \\
        \textbf{Steps to Follow:} \\
        \\
        0. \textbf{Extract Answers from both Responses:} \\
           - Read and both responses to identify the final answers provided. \\
           - If the responses provide different answers, make sure there are is no possible way that both responses can be correct. It must be the case that one response is correct and the other is incorrect or both are incorrect. \\
        \\
        1. \textbf{Step-by-Step Verification of Correctness:} \\
           - \textbf{For each response (Response A and Response B):} \\
             Carefully examine each step of the solution provided. Check the following: \\
             - \textbf{Mathematical accuracy:} Ensure all calculations, algebraic simplifications, and mathematical operations are correct. \\
             - \textbf{Logical consistency:} Verify that each step follows logically from the previous one and that the reasoning is sound. \\
             - \textbf{Completeness:} Make sure that all necessary steps are included to fully solve the problem and reach the final answer. \\
           \\
           While performing this step-by-step evaluation, refer to the \textbf{Additional Tips} section for helpful techniques to validate each response's accuracy. \\
           \textbf{Attention:} When checking the correctness of a single step, you should never first conclude the correctness of this step (for example, *"This step is incorrect because..."* is strictly forbidden). You should neutrally check this step, provide evidence about its correctness, and then finally draw a conclusion about the correctness of this step. In other words, you should first employ the techniques in \textbf{Additional Tips} to check the correctness of this step, and then draw a conclusion about the correctness of this step. \\
        \\
        2. \textbf{Final Conclusion:} \\
           - After completing the step-by-step verification for each response, sum up the information you have now, then finally determine whether each response’s answer is \textbf{correct} or \textbf{incorrect}. \\
           - Provide the final judgment for each response, the output should in-closed with the following tags: \\
             - If Response A’s answer is correct: \\
               \texttt{<resp\_a\_judge>Correct</resp\_a\_judge>} \\
             - If Response A’s answer is incorrect: \\
               \texttt{<resp\_a\_judge>Incorrect</resp\_a\_judge>} \\
             - If Response B’s answer is correct: \\
               \texttt{<resp\_b\_judge>Correct</resp\_b\_judge>} \\
             - If Response B’s answer is incorrect: \\
               \texttt{<resp\_b\_judge>Incorrect</resp\_b\_judge>} \\
           - \textbf{Note:} The responses A and response B can be either correct or incorrect, or both correct, or both incorrect. You should provide the final judgment for each response. There is no guarantee that at least one response is correct or incorrect. \\
        \\
        \textbf{Additional Tips:} \\
        \\
        - \textbf{Key Validation Techniques (to apply during Step 1):} \\
          - \textbf{Re-derive Key Parts of the Solution:} Independently calculate or derive crucial steps of the solution to verify their correctness. \\
          - \textbf{Verify Calculations:} Double-check all mathematical operations (e.g., addition, multiplication, division) to confirm accuracy. \\
          - \textbf{Compare Responses:} If needed, compare similar steps between Response A’s and Response B’s answers to identify discrepancies or inconsistencies. \\
        \\
        - \textbf{The final output format} should be as follows: \\
        \textbf{Final Judgment:} \\
        \textbf{Response A:} \texttt{<resp\_a\_judge>Correct/Incorrect</resp\_a\_judge>} \\
        \textbf{Response B:} \texttt{<resp\_b\_judge>Correct/Incorrect</resp\_b\_judge>} \\
        \\
        \textbf{Question:} \texttt{<question>} \green{\{question\}} \texttt{</question>} \\
        \textbf{Response A:} \texttt{<response\_a>} \green{\{response\_a\}} \texttt{</response\_a>} \\
        \textbf{Response B:} \texttt{<response\_b>} \green{\{response\_b\}} \texttt{</response\_b>} \\
        \bottomrule
    \end{tabular}
    \label{tab:math-judgment-task}
\end{table*}

\begin{table*}
    \centering
    \small
    \caption{Filtering criteria applied to the dataset to remove low-quality, proof-based, or multiple-choice problems.}
    \begin{tabular}{ll}
    \toprule
    \textbf{Filter Type}          & \textbf{Criteria}                                                                                  \\ 
    \midrule
    Bad Quality Problems          & Problems with messy formatting, OCR errors, or empty ground truth (\texttt{gt}).                  \\ 
    Equations in Ground Truth     & \texttt{gt} contains ``='' (indicating it might be an equation rather than a clear ground true).            \\ 
    Multiple Questions            & Problems with patterns indicating multiple sub-questions (\texttt{MULTI\_QUESTION}).              \\ 
    Yes/No Questions              & Solutions with patterns indicating yes/no, true/false (\texttt{YESNO\_QUESTIONS}).               \\ 
    Text Answers                  & Ground truth containing patterns indicating textual answers (\texttt{TEXT\_ANSWER}).             \\ 
    Proof Problems                & Problems with patterns indicating proof problems (\texttt{PROVE\_PATTERN}).                      \\ 
    Multiple Choice Questions     & Problems with patterns indicating multiple-choice questions (\texttt{MCQ\_OPTIONS}).             \\ 
    \bottomrule
    \end{tabular}

    \label{tab:math-problem-filter}
\end{table*}

\end{document}